\documentclass[10pt,twocolumn,final]{IEEEtran}
\usepackage[dvips]{graphicx}
\usepackage[para]{threeparttable}
\usepackage{amsmath}
\usepackage{caption2}
\usepackage[noadjust]{cite}
\ifCLASSINFOpdf
\else
\fi
\begin{document}
%
\title{Local Structure Matching Driven by Joint-Saliency-Structure Adaptive Kernel Regression}
%
%
%
%

\author{Binjie~Qin*,~Zhuangming~Shen,~Zien~Zhou,~Jiawei~Zhou, Jiuai Sun, Hui~Zhang, Mingxing Hu and Yisong Lv
\thanks{Manuscript received April 19, 2013; revised September 11, 2013; This work was supported in part by NSFC (61271320, 60872102 and 60402021), NBRPC (2010CB834300), China Scholarship Council and the small animal imaging project (06-545).}
\thanks{*Binjie Qin is with the School of Biomedical Engineering and Med-X Research Institute, Shanghai Jiao Tong University, Shanghai, 200240, China. During 2012 to 2013, he is also with Centre for Medical Image Computing, University College London, WC1E 6BT London, U.K. (e-mail: bjqin@sjtu.edu.cn).}
\thanks{Zhuangming Shen and Jiawei Zhou are with the School of Biomedical Engineering and Med-X Research Institute, Shanghai Jiao Tong University, Shanghai, 200240, China (e-mail: shenzhuangming@yahoo.com.cn; lenayui@gmail.com).}
\thanks{Zien Zhou is with Department of Radiology, Shanghai Renji Hospital, School of Medicine, Shanghai Jiaotong University (e-mail: zien.zhou@gmail.com).}
\thanks{Zhuangming Shen, Zien Zhou and Jiawei Zhou have contributed equally to this paper.}
\thanks{Jiuai Sun is with the Machine Vision Laboratory, University of the West of England, Bristol, BS16 1QY, U.K. (e-mail: jiuai2.sun@uwe.ac.uk).}
\thanks{Hui Zhang and Mingxing Hu are with Centre for Medical Image Computing, University College London, WC1E 6BT London, U.K. (e-mail: g.zhang@cs.ucl.ac.uk; mingxing.hu@ucl.ac.uk).}
\thanks{Yisong Lv is with the Department of Mathematics, Shanghai Jiao Tong University (e-mail: yslv@sjtu.edu.cn).}}

%
%

\markboth{Journal of \LaTeX\ Class Files,~Vol.~6, No.~1, January~2013}%
{Shell \MakeLowercase{\textit{et al.}}: Bare Advanced Demo of IEEEtran.cls for Journals}
%



\IEEEcompsoctitleabstractindextext{%
\begin{abstract}
For nonrigid image registration, matching the particular structures (or the outliers) that have missing correspondence and/or local large deformations, can be more difficult than matching the common structures with small deformations in the two images. Most existing works depend heavily on the outlier segmentation to remove the outlier effect in the registration. Moreover, these works do not handle simultaneously the missing correspondences and local large deformations. In this paper, we defined the nonrigid image registration as a local adaptive kernel regression which locally reconstruct the moving image's dense deformation vectors from the sparse deformation vectors in the multi-resolution block matching. The kernel function of the kernel regression adapts its shape and orientation to the reference image's structure to gather more deformation vector samples of the same structure for the iterative regression computation, whereby the moving image's local deformations could be compliant with the reference image's local structures. To estimate the local deformations around the outliers, we use joint saliency map that highlights the corresponding saliency structures (called Joint Saliency Structures, JSSs) in the two images to guide the dense deformation reconstruction by emphasizing those JSSs' sparse deformation vectors in the kernel regression. The experimental results demonstrate that by using local JSS adaptive kernel regression, the proposed method achieves almost the best performance in alignment of all challenging image pairs with outlier structures compared with other five state-of-the-art nonrigid registration algorithms.

\end{abstract}

}

\maketitle

\IEEEdisplaynotcompsoctitleabstractindextext

%
\IEEEpeerreviewmaketitle

\section{Introduction}
%
%

%
%
%
%
\IEEEPARstart
{N}onrigid image registration has attracted increasing attention at motion tracking, change detection, image segmentation and surface reconstruction in the computer vision and pattern recognition for the last decade \cite{1}-\cite{3}. The objective of nonrigid image registration is determining the local transformations that align every structures (or features) in one (moving) image with the corresponding structures in the another (reference) image. However, owing to the image content changes over a period of time and the different physical mechanisms of multimodal imaging sensors, some local structures presented in one image appear partially or even disappear completely in another image. Such local structures without one-to-one correspondence are closely intertwined with the strutures' local large deformations in the nonrigid image registration. The missing correspondences and local large deformations of structures are callled outliers in this paper. At present, obtaining an accurate and robust nonrigid image registration is still a challenging task for matching these local outlier strucutres with missing correspondences and/or the local large deformations. In the computer vision community, the nonrigid registration with the local large deformations is also knows as the large displacement optical flow problem \cite{4}. 

Generally, outliers mean the extreme observations substantially different from all other ones in the real data. For nonrigid image registration, the missing correspondences and local large deformations introduce the extreme observations which appear as the extreme local geometric and intensity differences between the two images to be registed. In geometric morphology, these outliers exhibit some large and complex structural discrepancies in the locally varying spatial contexts where the underlying different local structures could deform in substantially different ways. The desired nonrigid image registration should be able to match the moving image's local structures to the corresponding reference image's structures from these various local differences. Therefore, the \emph{local regression models} \cite{5}\cite{6}\cite{35} that are adept at handling the locally varying differences are necessary to account for these outliers, and it provides the rationale behind this work. Recently, Gerogiannis \emph{et al}.\cite{7} justify that the local Bayesian regression model are favorable to model local registration transformation compared with other interpolation based registration techniques.

Over the last several decades, many relevant works were proposed to match the local structures (or features) of the two images by minimizing the differences between the two images, which include feature-based, intensity-based, and hybrid methods \cite{1}\cite{2}. Feature-based \cite{8}\cite{9} approaches are local model based registration methods because they always use the local model of sparse image representation to select some corresponding features in the two images, and then directly match the local features by finding a geometrical transformation from these sparse feature correspondences. With the recent development of local feature detectors and descriptors (such as SIFT) \cite{4}\cite{10}, the feature-based methods have developed into the hybrid methods to use integrated local feature signatures \cite{11} in characterizing each voxel in the two images. Nevertheless, the computation of registration transformation is very sensitive to the ambiguity in finding local feature correspondences in different local spatial contexts with outliers. Moreover, most feature-based methods have not solved the outlier issues that have both missing correspondences and local large deformations.

By using the complete image data to recover dense correspondences at pixel-level precision, most intensity-based nonrigid registration approaches are regarded as global model based methods that are often formulated as \emph{global} energy minimization problems with the energy being composed of an regularization term and a similarity term \cite{12}\cite{13}. The relative weight of similarity term and regularization term can cause the well-known trade-off between the registration accuracy and the smoothness of the deformation field \cite{14}. In the presence of outliers, the accurate and plausible local structure matching does not exist using whole-intensity driven transformation model. The relative spatial regularization can either cause non-smooth and implausible distortion in these outlier regions or introduce over-smooth and inaccurate mapping artifacts between the whole images. This outlier problem can be solved by using a locally varying weight between regularization and image similarity \cite{15}-\cite{21}, creating artificial correspondences \cite{22}-\cite{27}, using cost-function masking \cite{28}-\cite{30}, or developing SIFT flow for large displacement \cite{4}. These approaches are largely dependent on the segmentation of outlier regions and give no full consideration to both the missing correspondences and local large deformations.

By successfully handling the locally varying differences in pattern recognition and machine learning, local kernel regression \cite{6}\cite{36} (or nonparametric regression) is regarded as an ideal local regression model to effectively reconstruct the desired local signal while suppressing the outlier and noise effects. The normalized convolution \cite{36} used by Suarez \emph{et al}. \cite{37}\cite{38} was the first application of local kernel regression in estimating dense deformation fields from sparse deformation fields. More recently, two works \cite{39}\cite{40} also utilized kernel regression to estimate registration transformations. Xing and Qiu \cite{42} proposed intensity-based nonparametric local smoothing to estimate local mapping transformation in a neighborhood. However, these methods do not consider the outlier problems in image registration.

To solve the outlier problem, we proposed the joint saliency map (JSM) to group the corresponding saliency structures (called Joint Saliency Structures, JSSs) in intensity-based similarity measure computation \cite{31}. The JSM has been proved to greatly improve the accuracy and robustness of rigid \cite{31}\cite{32} and nonrigid \cite{10}\cite{33}\cite{34} image registration with outliers. We further think, by reflecting the local structure correspondence, the JSM also could guide the local kernel regression for accurately estimating registration transformations in the nonrigid image registration with outliers. 

In this paper, by integrating the JSM into the kernel regression's local adaptive estimation, we propose a new JSS adaptive kernel regression to solve the outlier problem in the nonrigid registration. Specifically, with a moving window/kernel in kernel regression based nonrigid image registration, the output dense deformation vectors are locally estimated based on the specific weights of the sparse deformation vectors within the moving window. In the presence of outliers, the weights of the sparse deformation vectors for the outliers should be as small as possible to reduce the outlier effect on giving a distorted regression of dense deformations, while the JSSs and their underlying sparse deformation vectors should be emphasized with their weights being kept as high as possible to ensure the precision of the regression computation. Furthermore, the kernel function adapts its shape \cite{35},\cite{44}-\cite{47} and orientation to the reference image's local structure in order to gather more deformation vector samples of the same structure in the kernel regression, whereby the regression of local deformations can be locally compliant with the underlying local saliency structures without directing the deformation across the edges and corners of local structures.

To the best of our knowledge, this is the first work which propose the local JSS adaptive kernel regression to align the local structures by locally estimating the dense deformation fields of nonrigid image registration in the presence of outliers. An important contribution is that we use the JSM to guide local adaptive kernel regression for the accurate matching of local structures while suppressing outlier effects on the nonrigid image registration. The proposed method also makes the regression kernel not only locally adapt to the JSSs in the two images but also to the reference image's saliency structures. These two adaptivities enhance our local structure matching algorithm in achieving the accuracy and robustness of nonrigid registration of images with missing correspondences and local large deformations. The rest of this paper is organized as follows. Our algorithm is elaborated in Section 2 followed by experimental results in Section 3. The whole paper is concluded in Section 4.

\section{Methods}
\subsection{Coarse-to-fine Block Matching Scheme}
The algorithm proposed by us is built upon a coarse-to-fine iterative block matching scheme similar to the one in \cite{37}\cite{38}. The coarse-to-fine iterative framework is well known to deal with large deformation (or large displacement \cite{4}) in nonrigid image registration \cite{12}. Fig. 1 displays the whole coarse-to-fine iterative framework, where different levels have their own resolutions but the same procedure. At each level, the resulted global deformation is composed of initial deformation and current deformation. The initial deformation is obtained by resampling the global deformation from the previous level while the current deformation is the result of the current level involving the following two stages.

In the first stage, a discrete and sparse displacement field is created by maximizing local similarity measure for every pixel. The choice of local similarity measure has been studied in our previous work \cite{42}, where we employed cross-correlation (CC) instead of mutual information (MI) as the similarity measure when we matched two images at lower levels of the hierarchy so that the problem of MI's statistical consistency \cite{43} could be solved during the hierarchical subdivision. In this study, however, we only employ MI as the local similarity measure because we have restricted the block size of the lowest level of the hierarchy. Although block matching has many advantages in obtaining the deformations of an image, implementing only this algorithm is still insufficient to avoid the irregularity of transformations such as tearing, folding and distorting. Therefore, further reconstruction constraint is indispensable to be integrated into the registration procedure. To this end, a reconstruction procedure using local kernel regression is applied to this discrete and sparse displacement fields in the second stage. Details of this stage is described in Section 2.2 to Section 2.4. After the above-mentioned two stages for each level in this coarse-to-fine framework, a smooth and dense deformation field is iteratively achieved as the global deformation at the last and finest level.

\begin{figure}[!t]
\centering
\centerline{\includegraphics[width=3.4in]{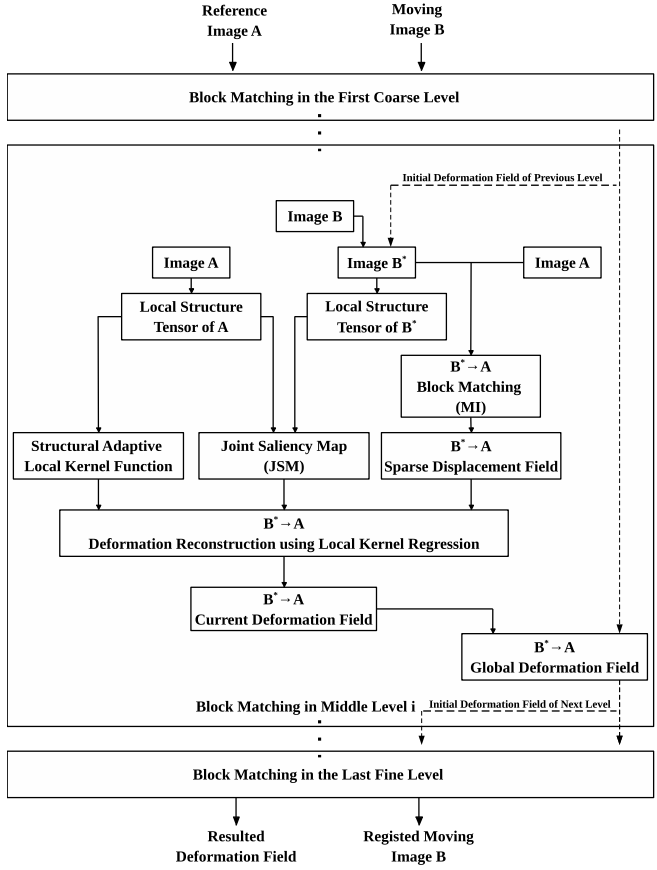}}
\caption{\small{\textbf{Flowchart of our algorithm within coarse-to-fine framework.} }}
\label{fig1.eps}
\end{figure}

\subsection{Kernel Regression Based Local Deformation Reconstruction}
Inspired by the successful applications in mordern image and video deblurring and super-resolution imaging \cite{35},\cite{44}-\cite{47}, we utilize kernel regression to reconstruct the dense deformation fields from the discrete and sparse displacement fields obtained through block matching. Suppose we have sparse and irregularly distributed deformation fields $\{\mathbf{y}_i,\mathbf{x}_i\}_{i=1}^{P}$ given in the form
\begin{equation}
\label{eq1}
\mathbf{y}_i = T(\mathbf{x}_i)+\mathbf{\varepsilon}_i,\quad \mathbf{x}_i\in\Omega,~~i=1,\cdots,P
\end{equation}
where $\mathbf{y}_i$ is a sparse displacement vector (response variable) at position (explanatory variable) $\mathbf{x}_i$, $T\left({\cdot}\right)$ describes the desired dense deformation field in the moving windows (kernel) $\Omega$ with independent and identically distributed zero mean noise ${{\varepsilon }_{i}}=\varepsilon \left( {{\mathbf{x}}_{i}} \right)$. In statistics, the function $T\left({\cdot}\right)$ is treated as a regression of $\mathbf{y}$ on $\mathbf{x}$, $T\left( \mathbf{x} \right)=E\left\{ \left. \mathbf{y} \right|\mathbf{x} \right\}$. In this way, the reconstruction of nonrigid deformation fields is from the field of the regression techniques.

Suppose the point of interest $\mathbf{x}$ to be reconstructed is near $\mathbf{x}_i$, then the regression of dense deformation field $T(\mathbf{x}_i)$ can be approximated by a local Taylor series expansion
\begin{equation}
\begin{split}
\label{eq2}
T(\mathbf{x}_i) & \approx ~T(\mathbf{x})+\{\nabla T(\mathbf{x})\}^T (\mathbf{x}_i-\mathbf{x}) \\
 & \quad +\frac{1}{2}(\mathbf{x}_i-\mathbf{x})^T\{\mathrm{Hessian}[T(\mathbf{x})]\}^T (\mathbf{x}_i-\mathbf{x})+\cdots \\
 & \approx \beta_0+\beta_1^T (\mathbf{x}_i-\mathbf{x}) \\
 & \quad + \beta_2^T \mathrm{vech}\{(\mathbf{x}_i-\mathbf{x})(\mathbf{x}_i-\mathbf{x})^T\}+\cdots
\end{split}
\end{equation}
where $\mathrm{vech}(\cdot)$ is a half-vectorization operator of a symmetric matrix and $\{\beta_0,\beta_1,\beta_2,\cdots,\beta_N\}$ are $N+1$ unknown parameters to be estimated. As in the 2D case, $\mathbf{\mathbf{x}}=[x_1,x_2]^T$, we can easily estimate the unknown parameters as
\begin{equation}
\begin{split}
\label{eq3}
& \beta_0 = T\left({\mathbf{x}}\right)  \\
& \beta_1 = [\frac{\partial T\left({\mathbf{x}}\right)}{\partial x_1},\frac{\partial T\left({\mathbf{x}}\right)}{\partial x_2}]^T  \\
& \beta_2 = \frac{1}{2} [\frac{\partial^2 T\left({\mathbf{x}}\right)}{\partial x_1^2},\frac{\partial^2 T\left({\mathbf{x}}\right)}{\partial x_1 \partial x_2},\frac{\partial^2 T\left({\mathbf{x}}\right)}{\partial x_2^2}]^T \\
& \cdots
\end{split}
\end{equation}
Since we have known the discrete displacement vectors $\{\mathbf{y}_i\}_{i=1}^P$, we can compute $\{\beta_n\}_{n=0}^N$ by finding the optimum solution of the following weighted least squares problem
\begin{equation}
\begin{split}
\label{eq4}
\mathrel{\mathop{\textstyle \min{}}\limits_{\scriptstyle \{\beta_0,\beta_1,\beta_2,\cdots \}}}  & \sum_{i=1}^{P} [\mathbf{y}_i-\beta_0-\beta_1^T (\mathbf{x}_i-\mathbf{x}) \\
& -\beta_2^T \mathrm{vech}\{(\mathbf{x}_i-\mathbf{x})(\mathbf{x}_i-\mathbf{x})^T\} \\
& -\cdots]^2 K_H(\mathbf{x}_i-\mathbf{x})
\end{split}
\end{equation}
where $K_H(\cdot)$ is a kernel function (see the detail in the next section), which not only smoothes the approximation but also penalizes distance away from $\mathbf{x}$.

In addition, if we assume $\mathbf{y}=[\mathbf{y}_1,\mathbf{y}_2,\cdots,\mathbf{y}_P]^T$, $\mathbf{b}=[\beta_0,\beta_1^T,\cdots,\beta_N^T]^T$, and $\mathbf{K}=\mathrm{diag}[K_H(\mathbf{x}_1-\mathbf{x}),K_H(\mathbf{x}_2-\mathbf{x}),\cdots,K_H(\mathbf{x}_p-\mathbf{x})]$, then we can rewrite the optimization problem in a matrix form
\begin{equation}
\begin{split}
\label{eq5}
\hat{\mathbf{b}}=\mathrel{\mathop{\textstyle \mathrm{argmin}}\limits_{\scriptstyle \mathbf{b}}} (\mathbf{y}-\mathbf{X}\mathbf{b})^T\mathbf{K}(\mathbf{y}-\mathbf{X}\mathbf{b})
\end{split}
\end{equation}
with
\begin{equation*}
\mathbf{X} =
    \begin{pmatrix}
     1      & (\mathbf{x_1-x}) & \mathrm{vech}^T \{(\mathbf{x_1-x})(\mathbf{x_1-x})^T \} & \cdots\\
     1      & (\mathbf{x_2-x}) & \mathrm{vech}^T \{(\mathbf{x_2-x})(\mathbf{x_2-x})^T \} & \cdots\\
     \vdots & \vdots           & \vdots                                                  & \vdots\\
     1      & (\mathbf{x_P-x}) & \mathrm{vech}^T \{(\mathbf{x_P-x})(\mathbf{x_P-x})^T \} & \cdots\\
    \end{pmatrix}
\end{equation*}
and the least-squares estimation solution can be expressed as
\begin{equation}
\begin{split}
\label{eq6}
\hat{\mathbf{b}}=(\mathbf{X}^T\mathbf{KX})^{-1}\mathbf{X}^T\mathbf{Ky}
\end{split}
\end{equation}
Because the zero-order Taylor series expansion known as the \emph{Nadaray}-\emph{Watson} estimator is sufficient to reconstruct the displacement vectors, the estimation of the deformation field at $\mathbf{x}$ has the form
\begin{equation}
\label{eq7}
\hat{T}(\mathbf{x})=\hat{\beta_0}=\frac{\sum_{i=1}^{P} K_H(\mathbf{x}_i-\mathbf{x})\mathbf{y}_i}{\sum_{i=1}^{P} K_H(\mathbf{x}_i-\mathbf{x})}
\end{equation}
Since images have outliers, it is reasonable to consider uncertainty for each pixel \cite{48}. Therefore, we add a weight (or certainty) function $c_i$ to equation (7)
\begin{equation}
\label{eq8}
\begin{split}
\hat{T}(\mathbf{x})=\hat{\beta_0} & = \frac{\sum_{i=1}^{P} K_H(\mathbf{x}_i-\mathbf{x})\cdot(\mathbf{y}_i\cdot c_i)}{\sum_{i=1}^{P} K_H(\mathbf{x}_i-\mathbf{x})\cdot c_i} \\
& = \frac{\mathbf{K} \otimes (\mathbf{y}\cdot \mathbf{c})}{\mathbf{K} \otimes \mathbf{c}} \\
\end{split}
\end{equation}
The last line of equation (8) can also be expressed in the form of normalized convolution \cite{36}, where $\otimes$ denotes convolution operation.

Fig. 2. illustrats the discrete displacement vectors reconstructed for every pixel in tumor resection region using local kernel regression. Because block matching results inherently contain incorrect matches, which are exacerbated by the outliers in the tumor resection region. As a result, the conflicts between neighboring displacement vectors (see the several red circles shown in Fig. 2(a)) widely exist in the discrete displacement field for the tumor region. Those displacement conflicts can easily introduce the topology change of structures, such as tearing and distorting. Fortunately, all the displacement vector conflicts are removed or smoothed by the local kernel regression in Fig. 2(b), where the displacement vectors having opposite directions fully disappear with the displacement magnitudes simultaneously being smoothed. Next, to match local structures in the presence of outliers, we design structural adaptive kernel functions and robust weighing schemes for the moving windows/kernels to further boost the accuracy and robustness of the local deformation reconstruction.
\begin{figure}[!t]
\centering
\centerline{\includegraphics[width=3.55in]{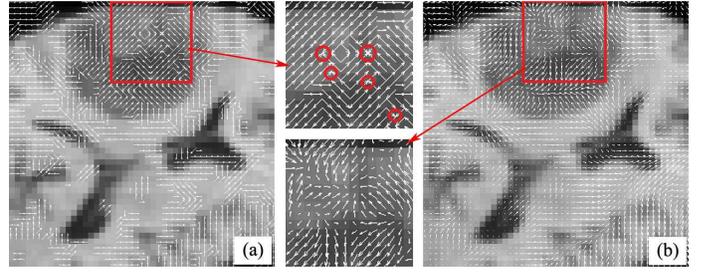}}
\caption{\small{\textbf{Application of kernel regression in deformation reconstruction.} (a) Discrete displacement field, (b) Reconstructed deformation field using kernel regression.}}
\label{fig2.eps}
\end{figure}

\subsection{Local structure-adaptive kernel function}
As a crucial component of local kernel regression, the shape of the kernel function (or moving window) determines the spatial distribution of samples which are gathered for the quality of the locally reconstructed signal. In principle, isotropic Gaussian kernels are mostly used as kernel functions in nonparametric regression. However, traditional isotropic Gaussian kernels are insufficient to cover more samples of the same modality along some specific orientations in signal reconstruction. Besides, Gaussian kernels' fixed scales and orientations can neither detect nor enhance edge structures. These factors easily cause diffusion across lines or edges of an image. To remedy these restrictions, Pham \emph{et al}. proposed an anisotropic kernel function to adapt its shape to the density of sampling \cite{46}. Afterwards, Takeda \emph{et al}. \cite{47} utilized gradient covariance matrices to construct steering kernels, which have been proved to possess the ability to capture the edges of an image and be extremely robust to noise and perturbations of the data.

In reconstruction of dense deformation field, the local kernel function extends along local structure orientation in the reference image so that it can gather more samples of sparse deformation vectors that correspond to the same saliency structures in the reference image. Besides, the kernel function contracts in the normal orientations of local saliency structures to prevent deformation diffusion across the edges between the different structures. Based on these schemes, it can effectively reduce the risk of changing the topology of the local saliency structures in the kernel regression. Considering that local structure tensor (LST) represents the anisotropic local saliency structure information of an image\cite{48}\cite{49}, we therefore design local structure-adaptive kernel functions using the LST information in the reference image. 

Before designing a local structure-adaptive kernel function at a certain pixel, we must compute LST in advance to grasp the local saliency structure around that pixel in 2D image. We assume that $I(\mathbf{x})$ denotes the intensity value at point $\mathbf{x}(x,y)$, the gradient information are expressed as $I_x=\frac{\partial I}{\partial x}$,~$I_y=\frac{\partial I}{\partial y}$, and $\nabla I(\mathbf{x})=[I_x~I_y]^T$ indicates the local gradient vectors, then the gradient structure tensor (GST) in 2D case can be described as
\begin{equation}
\label{eq9}
GST(\mathbf{x})=\nabla I(\mathbf{x})\nabla I(\mathbf{x})^T=
    \begin{pmatrix}
        I_x^2 & I_x I_y \\
        I_x I_y & I_y^2 \\
    \end{pmatrix}
\end{equation}
Generally, to integrate the surrounding structural information from neighborhoods, the GST is smoothed by a Gaussian filter to derive the LST. The scale $\sigma$ of the Gaussian filter is half the filter window size, namely $\sigma = 1.5$, because the size of the filter window in our experiments is $3\times 3$ pixels (or $3\times 3\times 3$ voxels for the 3D image). Therefore, we derive the LST from GST and perform a principal component analysis of the LST as follows:
\begin{equation}
\label{eq10}
LST(\mathbf{x})=G_\sigma\ast GST(\mathbf{x})=\lambda_u{\mathbf{u}}{\mathbf{u}}^T+\lambda_v{\mathbf{v}}{\mathbf{v}}^T
\end{equation}
where $\{\lambda_u,\lambda_v\}~(\lambda_u\geq\lambda_v) $ are the eigenvalues with the corresponding eigenvectors being $\left\{\mathbf{u,v}\right\}$. These eigenvectors contain the information about the local orientation distribution, in which pixel values change fastest along $\mathbf{u}$ direction while slowest along $\mathbf{v}$. Moreover, eigenvalues reflect both intensity variation in each direction and morphological information of a local region. For example in 2D image, $\lambda_u \approx \lambda_v \approx 0$ corresponds to a homogenous region with no measurable structure, $\lambda_u \gg \lambda_v \approx 0$ describes linear structure while $\lambda_u \geq \lambda_v \gg 0$ appears at corner structure.

With the above-mentioned LST computation in mind, we design the kernel functions to meet the following requirement: for regions containing obvious structural information such as histologic margins in a medical image, the kernels should be anisotropic with the regression computation being enhanced just along the main orientation and being suppressed along other orientations; for homogeneous regions without distinct structural information, the kernel should be isotropic with the regression computation being equal in all direction. We assume that $\mathbf{x_0}$ denotes a central position in 2D image, $\mathbf{x}$ is a position in its neighborhood, $\left\{\mathbf{u,v}\right\}$ are the eigenvectors of $LST(\mathbf{x}_0)$ and $\{\lambda_u,\lambda_v\}$ are the corresponding eigenvalues of $\left\{\mathbf{u,v}\right\}$, thus a local structure-adaptive Gaussian kernel in 2D case is designed as
\begin{equation}
\begin{split}
\label{eq11}
& a\left( \mathbf{x,}{{\mathbf{x}}_{\mathbf{0}}} \right)=\frac{1}{2\pi {{\sigma }_{u}}{{\sigma }_{v}}}\exp \left[ -\left( \frac{\mathbf{d}_{u}^{2}}{2{{\sigma }_{u}}}+\frac{\mathbf{d}_{v}^{2}}{2{{\sigma }_{v}}} \right) \right] \\
& {\mathbf{d}}_u=\langle \mathbf{d},\mathbf{u} \rangle, ~ {\mathbf{d}}_v=\langle \mathbf{d}, ~\mathbf{v} \rangle, ~\mathbf{d}=\mathbf{x}-\mathbf{x}_0 \\
\end{split}
\end{equation}
where $\mathbf{d}$ is the vector from $\mathbf{x_0}$ to $\mathbf{x}$, $\{{\mathbf{d}}_u,{\mathbf{d}}_v\}$ are the projections of the vector $\mathbf{d}$ on $\left\{\mathbf{u,v}\right\}$, and the directional scales of the Gaussian kernel $\{\sigma_u,\sigma_v\}$ are determined by the anisotropy $A$ as follows
\begin{equation}
\label{eq12}
\sigma_u = \frac{\alpha}{\alpha +A}\sigma_c, ~\sigma_v = \frac{\alpha +A}{\alpha}\sigma_c
\end{equation}
where $A={\left( {{\lambda }_{u}}-{{\lambda }_{v}} \right)}/{\left( {{\lambda }_{u}}+{{\lambda }_{v}} \right)}\;$. The two directional scales of the Gaussian kernel can be adjusted by the parameter $\alpha >0$ and the local scale $\sigma_c$. Specifically, $\alpha$ determines the eccentricity of the kernel while $\sigma_c$ affects the number of discrete vectors that contribute to the reconstruction of continuous deformation vectors. For the sake of simplicity, the local scale is set to half the neighborhood window size for each kernel according to our experience. We also set $\alpha = 0.5$ and $\sigma_c = 1.5$ because we utilize a $3\times 3$ pixel neighborhood window in our experiments. Similarily, the methodolgy of 3D LST and anisotropic kernel function computations is presented as an Appendix to this paper.
\begin{figure}[!t]
\centering
\centerline{\includegraphics[width=2.0in]{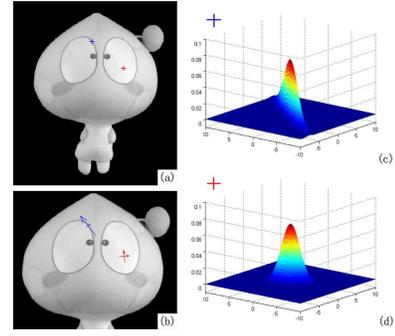}}
\caption{\small{\textbf{Gaussian kernels designed for different image local structures.} (a) Two labeled positions (red cross and blue one), (b) The scales and orientations of Gaussian Kernels in corresponding positions, (c) Gaussian kernel for the region with blue cross, (d) Gaussian kernel for the region with red cross.}}
\label{fig3.eps}
\end{figure}

Two kernel functions for two distinct image structures are displayed at the doll images in the Fig. 3, where the crosses indicate two different kinds of typical image structures (blue cross for border and red cross for homogeneous region). In Fig. 3, the two pairs of orthogonal vectors indicate the main axes of their corresponding Gaussian kernels (blue cross at Fig. 3(c) and red cross at Fig. 3(d)). The length of the vector is determined by the scale in the direction of the vector.
\begin{figure}[!t]
\centering
\centerline{\includegraphics[width=3.55in]{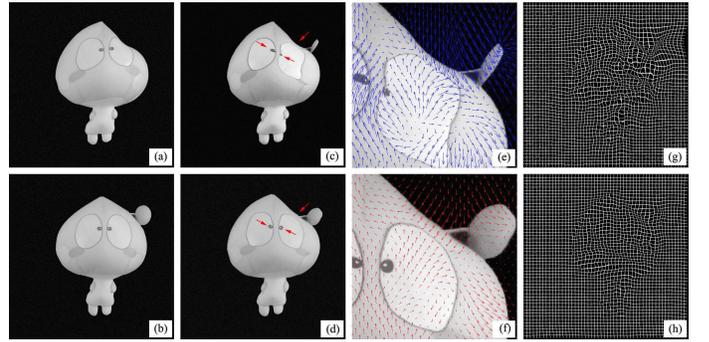}}
\caption{\small{\textbf{Comparison between using isotropic kernels and using local structure-adaptive kernels in kernel regression based deformation reconstruction.} (a)-(b) The reference and moving images, (c)-(d) Registered images using isotopic kernels and using local structure-adaptive kernels in kernel regression, (e)-(f) Local displacement vector fields for (c)-(d), (g)-(h) Global deformation mesh fields for (c)-(d).}}
\label{fig4.eps}
\end{figure}

Fig. 4 illustratively explain why we prefer local structure-adaptive kernel to isotropic kernel in our structure-adaptive kernel regression. The regions pointed by red arrows are small scale structures which have local large deformations. The directions of the displacement vectors (spaced every 5 pixels) in these small structures are inevitably conflicted with those of the large deformations of the neighboring structures. These defomration conflicts that are introduced by the opposite displacement vectors can easily cause tearing, folding or distorting of the local small scale structures. For example, the eyes in Fig. 4(c) display distortion owing to the conflict of the deformation directions displayed in Fig. 4(e). Comparatively, our local structure-adaptive kernels suppress the contributions of the displacement vectors which are not consistent in the structure orientation, the deformation conflicts are therefore removed in Fig. 4(f) with no distortion existing in the eyes at Fig. 4(d).

Fig. 4(h) demonstrates the overview of mesh deformation (10 pixels vertex spacing) in the local structure-adaptive kernel regression, which can produce the smooth adaptivity of local meshes deformation to the local anisotropic structures. Obviously, this local structure-adaptive kernel regression obtain smooth mesh boundaries which are consistent with the local structures' boundaries. However, isotropic kernel in kernel regression easily produce irregularly deformed local meshes which are not adaptive to the local structures, so that it is very difficult to identify boundaries of local structures from the non-smooth mesh deformation in Fig. 4(g).

\subsection{Robust Weight Mechanism using JSM}
In original kernel regression, the weight $c(\mathbf{x})$, between 0 and 1, specifies the reliability (or certainty) at $\mathbf{x}$ for the local estimation in a moving window/kernel. In modern local regression model, the weight function always describes the spatial dependence in the locally varying special context. With locally data-dependent weights, recently popularized and very effective image filter are developed in image and video processing field \cite{6}. For example, the Gaussian function of a residual error with an acceptable range of error $\sigma_r$ \cite{46} severs as a new weight function in the neighborhood of $\mathbf{x_0}$
\begin{equation}
\label{eq13}
c(\mathbf{x},\mathbf{x}_0)=\mathrm{exp} (-\frac{|I(\mathbf{x})-\hat{I}(\mathbf{x},\mathbf{x}_0)|^2}{2\sigma_r^2})
\end{equation}
where $I(\mathbf{x})$ denotes a measured intensity at position $\mathbf{x}$, and $\hat{I}(\mathbf{x},\mathbf{x}_0)$ is an estimated intensity at $\mathbf{x}$ using an initial polynomial expansion at $\mathbf{x}_0$.

Since the regions with salient structure information have real influence over locallly adaptive image processing based on kernel regression, E. Suarez \emph{et al}. \cite{37} proposed a weight scheme in the kernel regression of registration transformations by using the scalar measure of a local structure in reference image. However, for nonrigid image registration, the salient structural regions in reference image may introduce non-corresponding salient structural regions at same locations in moving image. Therefore, the method proposed by E. Suarez \emph{et al}. \cite{37} is most likely to assign wrong high weights to the salient structures that have missing correspondences. 

To avoid the above-mentioned mis-assignment and minimize the ourlier impact on the reconstruction of deformation field, we propose a robust weight mechanism by simultaneously considering the matching degree of local salient structures in the overlapping parts of the two images. In our previous work \cite{31}, it has been proved that the application of JSM is effective in tackling registration problems with outliers by emphatically grouping the JSSs into intensity-based similarity measure. Continuing the success of JSM, we further deploy the concept of JSM into our robust weight mechanism and pay more attention to the JSSs in the two images. Those JSSs and their incurring deformation should be emphatically treated in the local kernel regression to reconstruct the dense deformation fields from the sparse deformation fields that contain outliers.

With the general JSM-based weight scheme in mind, we should first construct a saliency map to indicate the local salient structure distribution in each image. Since LSTs mentioned in Section 2.3 contain sufficient structural information in a region, we can reasonably use them to reflect the saliency map of an image. Inspired by the center-surround mechanism \cite{50} which has defined the intensity-contrast-based visual saliency map, we define a saliency operator based on the contrast among neighboring structure tensors. This contrast emphasizes the dissimilarity or discrepancy between neighboring local structure tensors. Specifically, for a given point $\mathbf{x}_0$ and its neighborhood $\Omega$, the saliency value $S(\mathbf{x_0})$ at $\mathbf{x}_0$ in a salient map can be computed through
\begin{equation}
\label{eq14}
S(\mathbf{x_0})=\mathrm{avg}\sum_{\mathbf{x}\in\Omega}\|LST(\mathbf{x})-LST(\mathbf{x}_0)\|_D
\end{equation}
where $\|\cdot\|_D$ defines a distance metric describing the dissimilarity between two LSTs, which is detailed in the following section. The operator avg computes the average of the dissimilarities within the neighborhood $\Omega$ of $\mathbf{x}_0$. Traditional tensor similarity measures such as fractional anisotropy (FA) and cosine similarity measure are not appropriate for defining tensor-based saliency operator because they only compare either scales or orientations of two tensors. Fortunately, some improved tensor similarity measure computing both scale information and orientation information have been reported. In \cite{51}, H. Zhang \emph{et al}. introduced diffusion tensor metric, which is defined as
\begin{equation}
\label{eq15}
\|T_1-T_2\|_L=\sqrt{\frac{8\pi}{15}(\|T_1-T_2\|_C^2+\frac{1}{2} Tr^2(T_1-T_2))}
\end{equation}
where $\|T_1-T_2\|_C =\sqrt{Tr(T_1-T_2)^2}$ is the Euclidean distance between two tensors $\{T_1,T_2\}$, $Tr$ is the operator for computing the trace of matrix. Afterwards, H.~Zhang \emph{et al}. \cite{52} modified equation (15) to only focus on the anisotropic components between two tensors. The modified equation that is adopted at equation (14) can be expressed as
\begin{equation}
\label{eq16}
\|T_1-T_2\|_D=\sqrt{\frac{8\pi}{15}(\|T_1-T_2\|_C^2-\frac{1}{3} Tr^2(T_1-T_2))}
\end{equation}

In a saliency map, the saliency values are general representation of the local structure distribution in an image. Low saliency values always appear in the homogeneous and background regions while high saliency values are assigned to the edge regions of saliency structures owing to the highligted contrast among neighboring LSTs in these regions. After the two normalized salient maps were achieved to indicate the local structure distribution, JSM is ready to describe the matching degree between the two saliency maps at every pixel pair in the overlapping regions of the two images. Given a point $\mathbf{x}_R$ in the reference image and its corresponding point $\mathbf{x}_M$ in the moving image after initial transformation, their joint-saliency value in a JSM is defined as
\begin{equation}
\begin{split}
\label{eq17}
& \quad JS(\mathbf{x}_R,\mathbf{x}_M) \\
& =\min\{S_R(\mathbf{x}_R),S_M(\mathbf{x}_M)\}\frac{A\cdot B}{B+\|LST(\mathbf{x}_R)-LST(\mathbf{x}_M)\|_D}
\end{split}
\end{equation}
where $\{S_R(\cdot),S_M(\cdot)\}$ denote the saliency values in the saliency maps of the reference and the moving images. The empirical parameter $A$ and $B$ are used to normalize the JSM values into a final value between 0 and 1. In our experiments, $A=10$ and $B=\frac{1}{2}\max(\|LST(\mathbf{x}_R)-LST(\mathbf{x}_M)\|_D$. It should be noted that it may introduce a situation that both of the corresponding pixels are assigned high saliency values in the saliency maps, while their local variations of gradient orientations are in fact totally different. To avoid this situation, we also consider the dissimilarity measure between $LST(\mathbf{x}_R)$ and $LST(\mathbf{x}_M)$ at the denominator in equation (17).

\begin{figure}[!t]
\centering
\centerline{\includegraphics[width=2.5in]{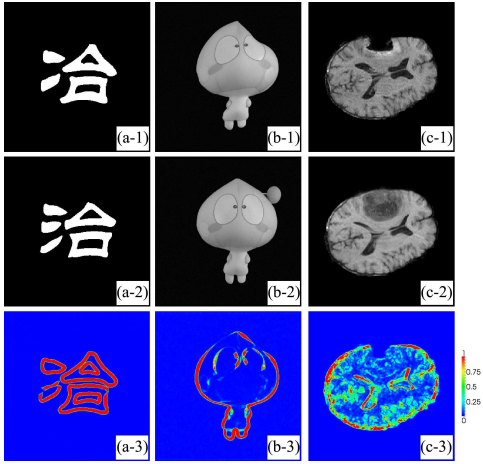}}
\caption{\small{\textbf{JSM Examples with colour scale representing different joint saliency values.} Each column shows one case. The reference images and the moving ones are shown at the top and middle rows. Their JSMs are displayed at the bottom rows where the red regions correspond to the higher joint saliency values while the blue regions correspond to the lower joint saliency values.}}
\label{fig5.eps}
\end{figure}

Fig. 5 shows some examples of normalized JSM with the colour scale representing different joint saliency values. The high joint saliency values being represented by red colour suggest that the underlying pixel pairs come from the JSSs. On the contrary, the regions with low JSM values are rendered in blue colour, which indicates that the underlying pixel pair originates from either homogeneous regions or outlier regions. The discrete displacement vectors in these red JSS regions are expected to contribute more to the kernel regression than the blue regions having low JSM values, this scheme is therefore called JSS adaptive kernel regression for nonrigid image registration.

\begin{figure}[!t]
\centering
\centerline{\includegraphics[width=3.55in]{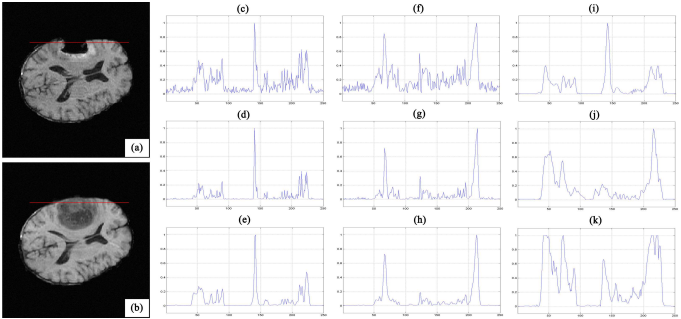}}
\caption{\small{\textbf{The reference and the moving images and their gradient magnitude, largest eigenvalue profiles for GST and LST, and the JSM magnitude.} (a)-(b) The reference and moving images, (c)-(e) Gradient magnitude profiles, largest eigenvalue profiles of GST and largest eigenvalue profiles of LST of the red line in (a), (f)-(h) Gradient magnitude profiles, largest eigenvalue profiles of GST and largest eigenvalue profiles of LST of the red line in (b), (i)-(j) Saliency value profiles of the red lines in (a)-(b), (k) JSM value profiles of the red lines in (a)-(b).}}
\label{fig6.eps}
\end{figure}

The JSM in our study mainly responds to the corresponding high-gradient edge pixels. However, it does not simply highlight the common image gradients in the two images. Fig. 6 presents the image gradient magnitude, the largest eigenvalues of GSTs and LSTs, the saliency values and the JSM values profiles of the same red line at the two images (Fig. 6(a)-(b)). For easy comparison, the range of ordinates in Fig. 6(c)-(k) are bound to [0,1]. As shown in Fig. 6, the image gradient features in Fig. 6(c) and Fig 6(f) are very sensitive to noise and do not agree with each other at each overlapping leocation. The noise sensitivity are gradually reduced by using the GSTs (Fig. 6(d) and Fig. 6(g)) and the LSTs (Fig. 6(e) and Fig. 6(h)). The saliency values of the two images in Fig. 6(i)-(j) are robust to noise due to their computing the regional contrast of LSTs through equation (14). Moreover, the structural image information in a large region is also comprehensively considered according to equation (14). As a result, the JSM values (Fig. 6(k)) computed through the saliency values can accurately preserve the JSSs in larger capture range with smaller variability than the image gradients.

Because of the outliers introduced by missing correspondence, local large deformation and incorrect block matching, the dense deformation fields can not be interpolated form the sparse displacement vectors in block matching. The JSS adaptive kernel regression is then used to reconstruct the dense deformation fields from the sparse displacement vectors, i.e., smooth the impact of outliers on the deformation reconstruction. Due to the expected deformation in the outlier region being consistent with its neighboring deformations, the JSM in the neighboring regions is used to assign different weights to the different displacements of the neighboring structures, only those neighboring deformations with high JSM values indicating the consistency in structure orientations are given high weights in kernel regression based deformation reconstruction.
\begin{figure}[!t]
\centering
\centerline{\includegraphics[width=3.5in]{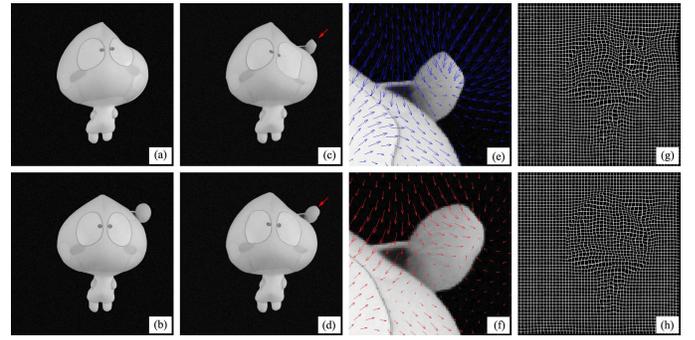}}
\caption{\small{\textbf{Comparison between using JSM-based robust weight mechanism and not using it in our method.} (a)-(b) The reference and moving image. (c)-(d) Registered images without and with using a JSM-based robust weight mechanism, (e)-(f) Local displacement vector fields of (c)-(d), (g)-(h) Global deformation mesh of (c)-(d).}}
\label{fig7.eps}
\end{figure}

Fig. 7 illustrates the improvements on the deformation field reconstruction after introducing the JSM-based robust weight mechanism in the local JSS adaptive kernel regression. The regions pointed by red arrows in Fig. 7(c)-(d) are outlier regions with missing correspondences and local large deformations. Without JSM-based robust weight mechanism, the converged displacement vectors (5 pixels spacing) from conflicting directions (See Fig. 7(e)) in the outlier region spread the distortion effect into the eye region (see Fig. 7(c)). Due to the JSM-based robust weight mechanism introducing weighted smoothing effect on the magnitude and direction of displacement vectors (see Fig. 7(f)), the moving image's eye distortion is removed with the outlier region's structure deformations being simultaneously matched to those of the reference image (see Fig. 7(f)). Compared with the deformation meshes (10 pixels vertex spacing) in Fig. 7(g), the deformation meshes at Fig. 7(h) display the overall smoothness improvement for the structural deformations at the outlier regions.

\section{Experimental Results}
In comparison with several state-of-the-art nonrigid registration approaches to validate the proposed algorithm on some challenging images with missing correspondences and local large deformations, we choose Advanced Normalized Tools (ANTs)\footnote{http://www.picsl.upenn.edu/ANTs} with elastic transformation and MI (AMI) \cite{53} due to the ANTs being placed the first in the EMPIRE10 challenge \cite{56} evaluating 34 total registration algorithms. We also include Diffeomorphic Demons with Diffusion-like regularization (DDD)\footnote{http://mipav.cit.nih.gov} \cite{54}, fast B-Spline with MI (BMI)\footnote{http://www.slicer.org}, AMI with cost-function Masking (AMM) and Large displacement Optical Flow (LOF) algorithms\footnote{http://www.seas.upenn.edu/\~{}katef/LDOF.html} \cite{4}. The parameters of our methods are: the number of pyramid levels is 5; the local similarity measure is mutual information. We set the parameters of AMI and AMM as: the histogram bin size is 32; the number of pyramid levels is 3; the iterations are set to $100\times100\times10$; the gradient step is 10; the default regularization is Gaussian filtering with a sigma of 3. The parameters of DDD method are set as follows: the variance of smoothing kernels is 2; the step scale is 1; the number of pyramid levels is 5; the number of iterations is 100. The parameters of the BMI method are selected as: the number of iterations is set to 100; the grid size is 15; the histogram bin size is 32; the spatial sample is 50000; the maximum deformation is 20. We choose the default parameters of the LOF method as: the tuning parameters for regularity constraint, the point correspondence constraint and the contraint on the gradient are 30, 300 and 5; downsampling factor is 0.95; the numbers of outer iterations and inner iterations are set to 10 and 5. With those parameters all the methods mentioned above achieve their best performances.

To evaluate the performance of the six competing methods, both registration accuracy and efficiency are estimated during the assessment. Validating the registration accuracy of binary image pairs is easy to recognize the badly-aligned regions by using the difference image between the reference and the registed moving images. However, the evaluation based on the difference images is not reliable for grayscale image registration \cite{55}. Due to the registration errors measured at densely distributed landmarks being considered as the standard for evaluating registration accuracy of grayscales images \cite{55}, we manually selected a large number of appropriate landmark pairs in the reference and the registered moving images to accurately evaluate the registration accuracy of the grayscale images. The selected landmarks fully exclude outlier features but identify some corresponding small scale saliency structures (having local large deformations) around the outlier regions. Therefore, the matching accuracies of six registration methods for the local structures with outliers are fully assessed by using the Mean Registration Error (MRE) and Standard Deviation (SD) in pixels between these selected landmarks. 

\begin{figure}[!t]
\centering
\centerline{\includegraphics[width=3.55in]{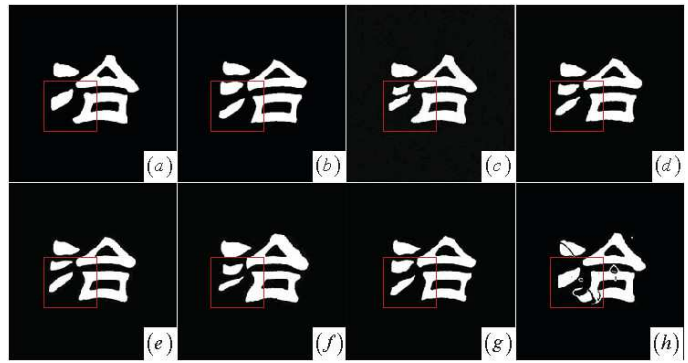}}
\caption{\small{\textbf{Chinese character image registration.} (a)-(b) The reference and moving images (c) the proposed method, (d) AMI, (e) DDD, (f) BMI, (g) AMM, (h) LOF.}}
\label{fig8.eps}
\end{figure}
\begin{figure}[!t]
\centering
\centerline{\includegraphics[width=3.4in]{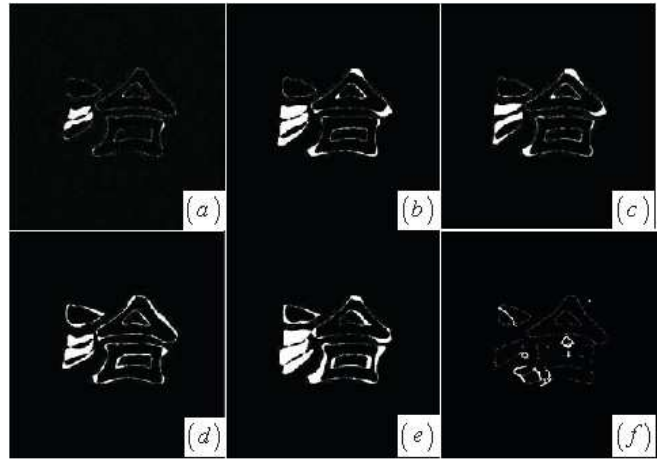}}
\caption{\small{\textbf{Difference images from the six Chinese character registration results.} (a) the proposed method, (b) AMI, (c)DDD, (d) BMI, (e) AMM, (f) LOF.}}
\label{fig9.eps}
\end{figure}

The reference and the moving binary images in the first experiment are two similar Chinese characters at Fig. 8(a)-(b). The two strokes in the left part of the reference image correspond to the upper left and lower left ones in the moving images, while the middle stroke (outlier) in the left part of moving image has missing correspondence (see the red rectangles in the Fig. 8). Moveover, the triangular and the rectangular openings at the right part of the character are narrowed. The local large deformations are apparent at the lower left corner of the rectangular region and the low right corner of the triangular region. Besides, the lower left stroke from southwest to northeast is lengthened with counter-clockwise rotation. Fig. 8(c)-(g) are the registered results of our approach, AMI, DDD, BMI, AMM and LOF. The strokes at the left part and the corner regions of the two openings are locally deformed more or less in the six registered images. Though the constrained cost-function masking has masked the middle stroke in moving image, AMM method perform poorly in matching local structures with local large deformations in Fig. 8(g). The LOF method changed the topology of moving image by fully removing the middle stroke in the left part and introducing some artifacts in the character at Fig. 8(h). Overall, the proposed method (see Fig. 8(c) has produced the best structure deformation among the six registered results.

The performance of our proposed approach could be clearly validated from the difference images (see Fig. 9) of the six registration results, where the white regions represent the discrepancies between the reference image and the registered moving image. Althogh the difference image of LOF method has smallest white regions among the six methods, the LOF method also introduce distinct artifacts and topology change in the registered Chinese character. Without robust scheme tackling missing correspondence and local large deformation simultaneously, the AMM method (Fig. 9(e)) even performs a little worse than the AMI method (Fig. 9(b)). The white regions in the difference image (Fig. 9(a)) of our approach are the least among the other five results which preserve the topology of the middle stroke. It validates that our proposed method match well the moving image's local structures into the corresponding structures at the reference image. 
\begin{figure}[!t]
\centering
\centerline{\includegraphics[width=3.55in]{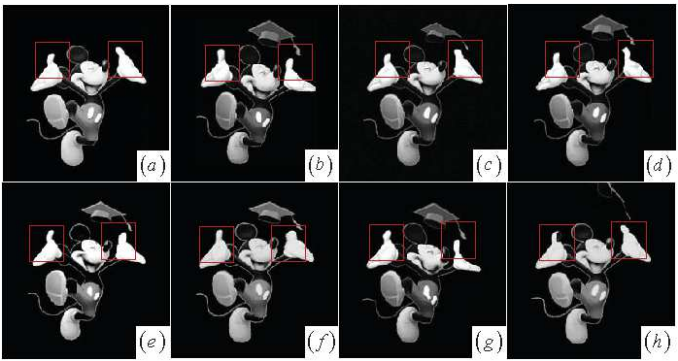}}
\caption{\small{\textbf{Mickey image registration.} (a)-(b) The reference and moving images, (c) the proposed method, (d) AMI, (e) DDD, (f) BMI, (g) AMM, (h) LOF.}}
\label{fig10.eps}
\end{figure}

The second experiment involves aligning two grayscale Mickey images, which have the outlier doctoral hat shown in the moving image (Fig. 10(b)). Moreover, the local structure's large deformations occur in Mickey's left thumb, left hand, right thumb (see the red rectangles in the Fig. 10), right shoe and right button in Mickey's belly. Consequently, validating the performance of registration methods is largely dependent on the deformed results of these structures. Fig. 10(c)-(h) show the registered results of the six methods, where our proposed approach outperforms the other five methods by perfectly deforming those local structures to desired positions. In contrast, the morphologies of Mickey's left hand in Fig. 10(d) and (g) are changed by AMI and AMM methods. The Mickey's left thumb, right shoe and right button in Fig. 10(e) have not changed by DDD method. The Mickey's left thumb has not deformed and the left palm become thicker in Fig. 10(f) after BMI registration. In this case, though with the constrained cost-function masking for the doctoral hat, AMM method has no improvement in matching local structures (Fig. 10(g)). The LOF method introduces an undesired artifact in the right thumb of the Mickey (see the left red rectangle in the Fig. 10(h)).
\begin{figure}[!t]
\centering
\centerline{\includegraphics[width=3.55in]{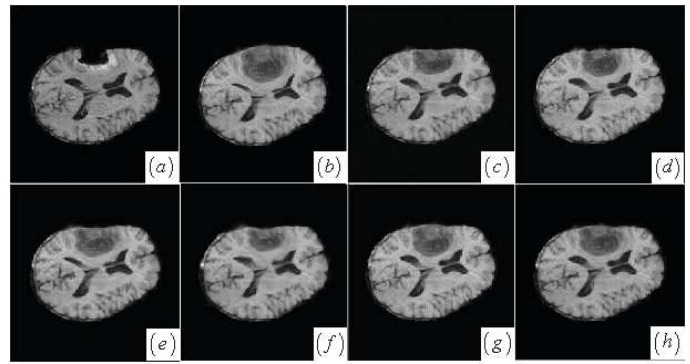}}
\caption{\small{\textbf{Brain tumor resection image registration.} (a)-(b) The reference and moving images, (c) the proposed method, (d) AMI, (e) DDD, (f) BMI, (g) AMM, (h) LOF.}}
\label{fig11.eps}
\end{figure}
\begin{figure}[!t]
\centering
\centerline{\includegraphics[width=3.55in]{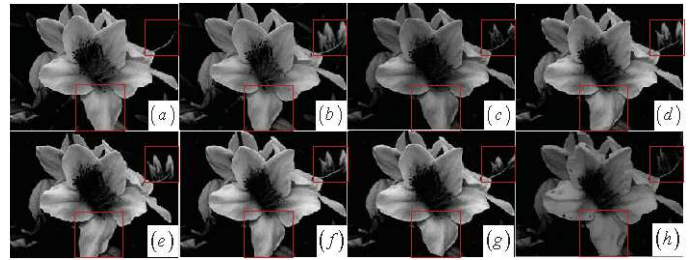}}
\caption{\small{\textbf{Flower image registration.} (a)-(b) The reference and moving images, (c) the proposed method, (d) AMI, (e) DDD, (f) BMI, (g) AMM, (h) LOF.}}
\label{fig12.eps}
\end{figure}

Another grayscale image registration is matching pre- and post-operative brain tumor images. Brain tissue severely suppressed by tumor in the preoperative image (Fig. 11(b)) expands after tumor resection. Tumor resection not only brings missing correspondences into the tumor region of the post-operative image (Fig. 11(a)) but also incurs local large deformations that are caused by brain shift. A successful registered result of this case should properly deform pre-operative brain tissue according to the post-operative image regardless of tumor resection. Visual inspection has revealed that the proposed, AMI, AMM and LOF methods (see Fig. 11(c)-(d),(g)) apparently perform better than the DDD and BMI methods (see Fig. 11(e)-(f)) because the local brain deformation resulted from the latter two methods is either insufficient or somewhat excessive. This visual valuation is further confirmed by validating landmark-based registration error in the following section.

A more challenging flower image registration is displayed in Fig. 12, where the small scale stamen filament in the right part of the image is largely deformed while some buds behind the stamen filament of the moving image being disappeared in the reference image (see the top red rectangles in the Fig. 12). Matching small scale structures that have large deformations and missing correspondence is very difficult for nonrigid image registration. In the case of Fig. 12, an ideal registration algorithm should accurately match not only the large petal (see the bottom red rectangles) but also the small scale stamen filament while simultaneously keeping reasonable local deformation consistency in the spatial context. Among these six methods, only the proposed approach aligns not only the small scale stamen filament but also the large scale petal by computing their resonable and accurate deformations. The AMI (Fig. 12(d)) and AMM (Fig. 12(g)) methods achieve better registration performances than the DDD and BMI methods (Fig. 12(e)-(f)). The LOF (Fig. 12(h)) method gets good deformations for most structures, but also apparently intruduce some artifacts in the petals of the registered moving image.

\setlength{\tabcolsep}{2pt}
\begin{center}
\begin{table}
\centering
\renewcommand{\arraystretch}{0.7}
\caption{\small{\textbf{Registration errors (Mean+SD) of the six methods for the three grayscale image registrations} }}
\begin{tabular*}{0.5\textwidth}{@{\extracolsep{\fill}}llllll}
\hline
proposed    &AMI            &DDD            &BMI             &AMM      &LOF\\
\hline
1.27$\pm$3.09  &1.56$\pm$3.38    &6.07$\pm$5.93    &3.23$\pm$5.51   &12.03$\pm$14.80  &3.55$\pm$3.21\\
0.97$\pm$1.91  &0.95$\pm$1.48    &1.60$\pm$3.08    &1.15$\pm$1.89   &1.16$\pm$1.99  &1.97$\pm$1.93\\
1.14$\pm$2.96   &1.69$\pm$3.49     &8.38$\pm$8.82    &4.42$\pm$5.65   &1.55$\pm$3.49  &2.72$\pm$2.35\\
\hline
\end{tabular*}
\end{table}
\end{center}
The MRE and SD of the manually selected landmarks for the six methods in the three grayscale image registration are listed in Tab. 1. The proposed method for the Mickey image achieves the smallest registration error of $1.27\pm3.09$ pixels while the registration errors of AMI, DDD, BMI, AMM and LOF methods are greater than or equal to $1.56\pm3.38$ pixels. Compared with other method, the proposed method and AMI have achieved sub-pixel registration accuracy for the brain tumor resection images with the registration errors of $0.97\pm1.91$ and $0.95\pm1.48$ pixels, respectively. As for the flower image, the proposed method gets the smallest registration error of $1.14\pm2.96$ pixels, while the registration errors of other five methods are greater than or equal to $1.69\pm3.49$ pixels. In average, the proposed method maintains almost the best performance in comparison with other five methods. Although the orignial AMI method in the brain tumor resection image registration has a slight advantage over the proposed method, using cost-function masking makes AMM method worse than the proposed method in matching locally deformed structures. Simply setting the brain-tumor-region's value to zero by cost-function masking is not enough to accurately match local salient structures with missing correspondences and local large deformations. Due to the LOF method only considering the effect of large displacments, its performance is not desired for the nonrigid registration with both missing correpondences and local large deformations.

Table 2 lists the computation time needed for the six algorithms at the four image registration, where the image resolution of case 1-3 is 372$\times$392 pixels, and that of case 4 is 384$\times$288 pixels. All the six methods are operated on a PC of Pentium(R) Dual-Core 3.20 GHz CPU with 2 GB memory.
\begin{center}
\begin{threeparttable}
\centering
\renewcommand{\arraystretch}{0.8}
\caption{\small{\textbf{Computation runtime in seconds for the six registration methods} (Pentium(R) Dual-Core 3.20 GHz RAM 2.0 GB)}}
\begin{tabular*}{0.5\textwidth}{@{\extracolsep{\fill}}ccccccc}
\hline
Cases   &proposed   &AMI  &DDD  &BMI  &AMM &LOF  \\
\hline
1        &48             &38             &12         &31    &38   &112\\
2        &49             &56             &13         &42    &45   &144\\
3        &48             &21             &12         &43    &20   &113\\
4        &36             &21             &10         &75    &7    &168\\
\hline
\end{tabular*}
\end{threeparttable}
\end{center}

\section{Conclusion}
In this paper, considering the outlier structures that have missing correspondences and local large deformations in nonrigid image registration, we use JSS adaptive kernel regression to reconstruct dense deformation field (for the moving image) from the sparse deformation field hierarchically computed from multi-resolution block matching. Specifically, the proposed local JSS adaptive kernel regression implements two local adaptivities into the underlying saliency structures in the reference images and the JSSs in the two images to accurately and robustly match corresponding local structures with missing correspondence and local large deformation.

Nevertheless, there are still some issues that need to be further addressed in our future work. First, the local scale mentioned in Section 2.3 is set as a constant value for the sake of simplicity at each anisotropic Gaussian kernel. As shown in some registered results, however, the deformations of some small scale structures (such as the eyes in the doll image at Fig. 8) are affected somewhat badly by the deformations of the surrounding large scale structures. This situation is caused by the unchangeable local scales. Indeed, the local scale (the width of kernel) controls the number of discrete displacement vectors contributing to the reconstruction of dense deformation vectors from sparse displacement vectors. Therefore, the choice of local scale significantly affects the final registration result. A self-adjustable local scale according to the local structure properties is expected to automatically adjust the number of discrete displacement vectors participating in the deformation reconstruction. To our knowledge, almost no attention has been paid on the self-adjustable local scale for nonrigid image registration during the last decade.

Second, we could also improve the proposed algorithm by replacing the block matching with other feature-based or hybrid nonrigid registration algorithms so that we can accurately initialize deformation field estimation and apply the proposed method to match multimodal images. As for the diffeomorphism of nonrigid image registration \cite{53}\cite{54}\cite{56}, the missing correspondences and local large deformations make it unrealistic for nonrigid registration method enforcing the diffeomorphic local structure matching. However, we could initialize our method with some diffeomorphic registration algorithms to find the local structures' optimal diffeomorphic matching for the subsequent JSS adaptive kernel regression. At last, further researches are required to reduce the computational cost of our approach. At present, more than half of the total running time for our proposed method is spent on the local adaptive Gaussian kernel reconstruction and subsequent adaptive kernel regression at every pixel. We should design fast method \cite{57} to estimate discrete local structure-adaptive Gaussian kernel and implement structural adaptive kernel regression at every pixel. All these improvements could prompt us to make further contributions to the nonrigid image registration.


%

\appendices
\section{3D local structure tensor and anisotropic kernel function}
The 3D LST can be computed for every point $\mathbf{x}(x,y,z)$ in 3D image as follows. With the gradient being computed as $I_x=\frac{\partial I}{\partial x}$,~$I_y=\frac{\partial I}{\partial y}$,~$I_z=\frac{\partial I}{\partial z}$ and the $\nabla I(\mathbf{x})=[I_x~I_y~I_z]^T$ indicating the local gradient vectors, the 3D GST is estimated as
\begin{equation}
\label{eq18}
GST(\mathbf{x})=\nabla I(\mathbf{x})\nabla I(\mathbf{x})^T=
    \begin{pmatrix}
        I_x^2 & I_x I_y & I_x I_z \\
        I_x I_y & I_y^2 & I_y I_z\\
        I_x I_z & I_y I_z & I_z^2\\  
    \end{pmatrix} 
\end{equation}
while the smoothed 3D LST being
\begin{equation}
\label{eq19}
LST(\mathbf{x})=G_\sigma\ast GST(\mathbf{x})=\lambda_u{\mathbf{u}}{\mathbf{u}}^T+\lambda_v{\mathbf{v}}{\mathbf{v}}^T+\lambda_w{\mathbf{w}}{\mathbf{w}}^T
\end{equation}
where $\{\lambda_u,\lambda_v,\lambda_w\}~(\lambda_u\geq\lambda_v\geq\lambda_w) $ are the eigenvalues and their corresponding eigenvectors are denoted as $\left\{\mathbf{u,v,w}\right\}$. The voxel values change fastest along $\mathbf{u}$ direction while slowest along $\mathbf{w}$ direction.

To derive the 3D kernel function at $\mathbf{x_0}$, we express the anisotropic 3D Gaussian function $a\left( \mathbf{x,}{{\mathbf{x}}_{\mathbf{0}}} \right)$ using the eigenvalues and corresponding eigenvectors of 3D LST. 
\begin{equation}
\begin{split}
\label{eq20}
& a\left( \mathbf{x,}{{\mathbf{x}}_{\mathbf{0}}} \right)=\frac{1}{\sqrt[]{8\pi^3} {{\sigma }_{u}}{{\sigma }_{v}}{{\sigma }_{w}}}\exp \left[ -\left( \frac{\mathbf{d}_{u}^{2}}{2{{\sigma }_{u}}}+\frac{\mathbf{d}_{v}^{2}}{2{{\sigma }_{v}}}+\frac{\mathbf{d}_{w}^{2}}{2{{\sigma }_{w}}} \right) \right] \\
& {\mathbf{d}}_u=\langle \mathbf{d},\mathbf{u} \rangle , ~ {\mathbf{d}}_v=\langle \mathbf{d},\mathbf{v} \rangle,~ {\mathbf{d}}_w=\langle \mathbf{d},\mathbf{w} \rangle,~\mathbf{d}=\mathbf{x}-\mathbf{x}_0 \\
\end{split}
\end{equation}
where $\mathbf{d}$ is the vector from $\mathbf{x_0}$ to $\mathbf{x}$, $\{{\mathbf{d}}_u,{\mathbf{d}}_v,{\mathbf{d}}_w\}$ are the projections of the vector $\mathbf{d}$ on $\left\{\mathbf{u,v,w}\right\}$. In order to estimate the directional scales of the 3D anisotropic Gaussian kernel \cite{58}, we first compute the anisotropy $\{a_{vw},a_{uw}\}$ of 3D LST with
\begin{equation}
\label{eq21}
a_{vw} = \frac{\lambda_v-\lambda_w}{\lambda_u+\lambda_v+\lambda_w} ,~a_{uw} = \frac{\lambda_u-\lambda_w}{\lambda_u+\lambda_v+\lambda_w}
\end{equation}
We further estimate the spatial dependent corner strength $C(\mathbf{x_0})$ as
\begin{equation}
\label{eq22}
C\left(\mathbf{x_0}\right) = \left(1-a_{vw}-a_{uw}\right)|\nabla I(\mathbf{x_0})|^2
\end{equation}
where $|\nabla I(\mathbf{x_0})|^2$ is the local gradient strength at the point $\mathbf{x_0}$. Therefore, we obtaint the three directional scales of the 3D Gaussian kernel as
\begin{equation}
\begin{split}
\label{eq23}
& \sigma_u = \frac{\sigma_c(1-a_{vw}-a_{uw})}{1+C(\mathbf{x_0})} \\
& \sigma_v = \frac{\sigma_c(1-2a_{vw})}{1+C(\mathbf{x_0})},~\sigma_w = \frac{\sigma_c}{1+C(\mathbf{x_0})} \\
\end{split}
\end{equation}


\section*{Acknowledgment}
The authors would like to thank Simon K. Warfield for providing MR brain images, E. Suarez-Santana for opening the source code in ITK project and Katerina Fragkiadaki for opening the souce code of large displacement optical flow. The authors thank the open source ANTs project, MIPAV, Diffeomorphic Demons, 3D Slicer and ITK project.

\ifCLASSOPTIONcaptionsoff
  \newpage
\fi

\end{document}